\documentclass{article}
\usepackage{spconf,amsmath,graphicx}
\usepackage{mathtools}
\usepackage{url}
\usepackage{amsmath}
\usepackage{multirow}

\DeclareMathOperator*{\argmin}{arg\,min}
\title{Novel tile segmentation scheme for omnidirectional video}
\name{Jisheng Li, Ziyu Wen, Sihan Li, Yikai Zhao, Bichuan Guo, Jiangtao Wen}
\address{Tsinghua University\\ Beijing, 100084, China}
\begin{document}
\ninept
\maketitle
\begin{abstract}

Regular omnidirectional video encoding technics use map projection to flatten a scene 
from a spherical shape into one or several 2D shapes.
Common projection methods including equirectangular and cubic projection
have varying levels of interpolation that create a large number of non-information-carrying 
pixels that lead to wasted bitrate. 
In this paper, we propose a tile based omnidirectional video segmentation scheme
which can save up to 28\% of pixel area and 20\% of BD-rate averagely compared to the traditional 
equirectangular projection based approach.
\end{abstract}
\section{Introduction}
\label{sec:intro}

The ever wider application of virtual reality  (VR) requires that 
omnidirectional video for consumption on VR devices be efficiently captured, encoded
and transmitted.

Various types of equipment for  recording omnidirectional video has already been developed.
Most of these capturing equipments use two fish-eye cameras or multiple wide angle 
cameras to capture the scene and then stitch frames from these cameras together. In this process,
an omnidirectional picture well represented using a spherical surface is created, and then
mapped into a flat plane for subsequent encoding and then transmitted. 
A number of mapping methods from the spherical to flat surfaces 
have been proposed, such as the technique in \cite{snyder1997flattening} \cite{chen1995quicktime} \cite{szeliski1997creating},
with various distortions created during the process \cite{zorin1995correction}.

In this paper, we propose a tile-based segmentation and projection scheme. It 
attempts to create an approximate equal-area mapping and projection with 
compression-friendly shapes with less wasteful pixels.  
Our scheme can save up to 28\% of pixel area compare to 
equirectangular projection and 40\% of pixel area compare to 
cubic \cite{ng2005data} projection at the same minimum sampling density
in the original spherical surface.

The paper is organized as follows: In section 2, we review related work and 
the quality metric. Section 3 describes the proposed tile segmentation scheme 
in details.  Experimental results are shown in section 4 while section 5 is the conclusion.  
\begin{figure}[htb]
\begin{minipage}[b]{0.475\linewidth}
  \centering
  \centerline{\includegraphics[width=4.0cm]{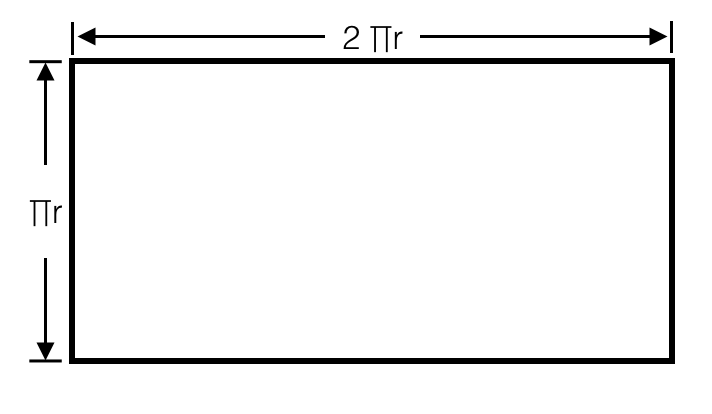}}
  \centerline{(a) Equirectangular (157\%)}
\end{minipage}
\hfill
\begin{minipage}[b]{0.475\linewidth}
  \centering
  \centerline{\includegraphics[width=4.0cm]{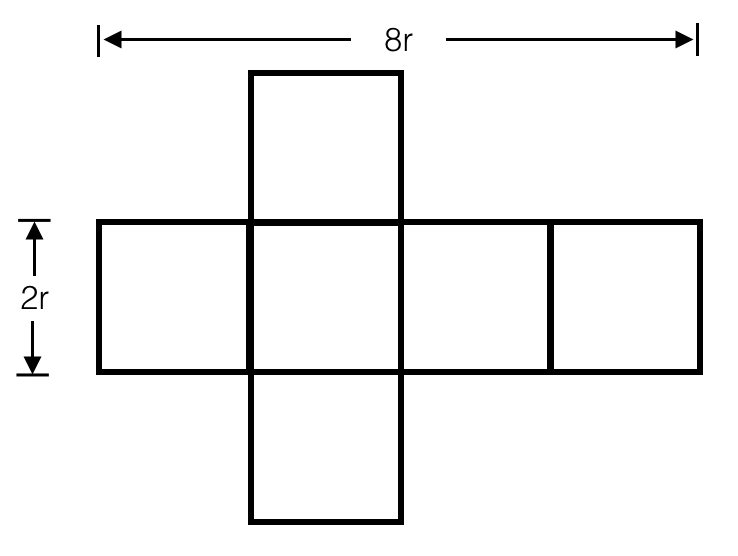}}
  \centerline{(b) Cubic (190\%)}\medskip
\end{minipage}

\begin{minipage}[b]{0.475\linewidth}
  \centering
  \centerline{\includegraphics[width=4.0cm]{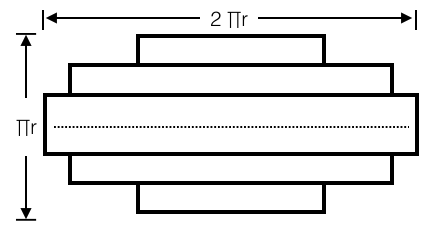}}
  \centerline{(c) Tile (123\%)}\medskip
\end{minipage}
\hfill
\begin{minipage}[b]{0.475\linewidth}
  \centering
  \centerline{\includegraphics[width=4.0cm]{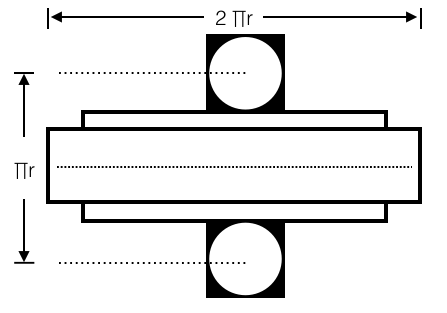}}
  \centerline{(d) Proposed Tile (113\%)}\medskip
\end{minipage}
\caption{Comparison of map projections under same minimum sample density on the sphere
(with percentage of original sphere area in the brackets). 
(c) A tile segmentation scheme with equal division on latitude;
(d) Our proposed 5-tiles scheme}
\label{fig:four}
\end{figure}

\begin{figure}[htb]
\begin{minipage}[b]{1.0\linewidth}
  \centering
  \centerline{\includegraphics[width=8cm]{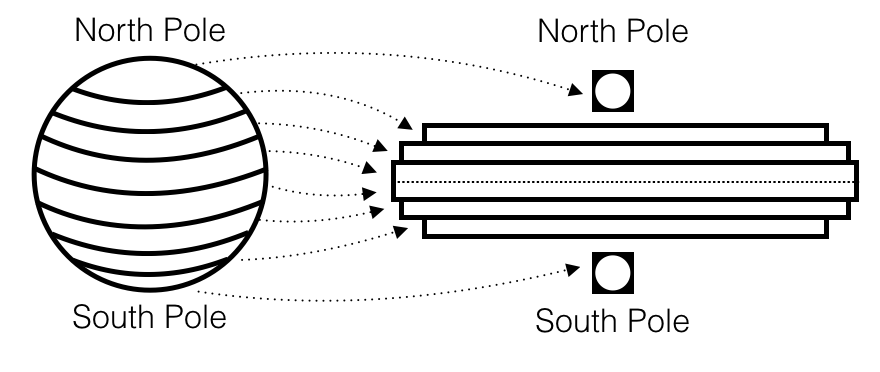}}
  \centerline{(a) A tile segmentation example}\medskip
\end{minipage}
\hfill

\begin{minipage}[b]{1\linewidth}
  \centering
  \centerline{\includegraphics[width=8cm]{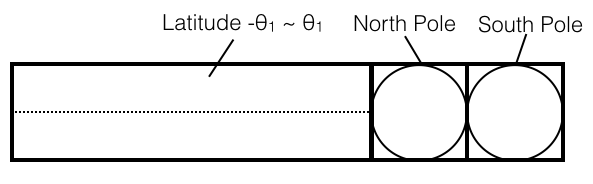}}
  \centerline{(b) A rearrangement layout for 3-tiles}\medskip
\end{minipage}
\hfill
\begin{minipage}[b]{1\linewidth}
  \centering
  \centerline{\includegraphics[width=8cm]{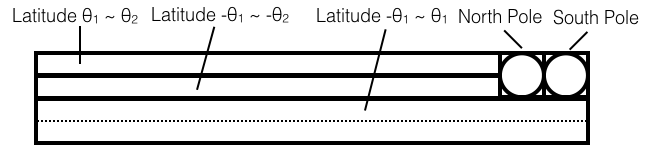}}
  \centerline{(c) A rearrangement layout for 5-tiles}\medskip
\end{minipage}

\caption{Tile segmentation and rearrangement illustration}
\label{fig:seg}
\end{figure}
\section{Related work}
\label{sec:r_work}

Many early techniques for capturing omnidirectional video 
\cite{tovsic2009low} \cite{bauermann2006h} used a camera and a parabolic mirror. 
The camera captures images reflected by the parabolic mirror, with 
the vertical field of view limited in 90$^{\circ}$.

Lately, higher quality omnidirectional video are captured with multiple high definition 
cameras with the captured frames stitched \cite{rondao2012interactive}. When projecting the image to a flat 2D 
plane, equirectangular projection was widely used. Despite of its ease of use,
because the areas in the north and south poles of the sphere are stretched 
more severely than areas near the equator, overall, a net increase of 57\% 
of the original sphere area was introduced wastefully, leading to a resultant 
increase in bitrate. Furthermore, due to the distortions introduced by
the stretching, especially near the two poles, predictive coding tools in video encoders 
fail more easily, causing further reduction in coding efficiency.  

In \cite{ng2005data}, Ng et al. proposed a method to project a sphere onto a cube, referred 
as cubic projection subsequently. Although the distortion introduced by projection is mitigated in
cubic projection, the total area of the 6 sides has to be 190\% of the original sphere area 
so that the central of each surface has enough sample density,
as is shown in Fig.\ref{fig:four}. 
Later in \cite{fu2009rhombic}, Fu et al. proposed a rhombic dodecahedron map projection scheme, 
which divided a sphere into 12 rhombs then reshape to squares and rearrange into a rectangular.

In \cite{yu2015content}, Yu et al. divided the video vertically into tiles 
and resized the tiles according to the latitudes or user preferences (e.g. areas near the poles might be less likely
to be requested by the viewer). Even though this scheme solved many of the issues caused by stretch related 
distortions and increase in the total number of pixels after mapping, it did not fully address the issue of 
preserving effectiveness of predictive coding tools in video encoders. 

There are relatively few studies on methods for evaluating video coding quality 
for omnidirectional video. Such evaluation is a challenge because after the input 
omnidirectional video is projected onto a 2D plane and then coded, many of the pixels in the resulted
video may be interpolated and should not carry the same ``weight'' as in regular PSNR calculation. 
In \cite{yu2015framework}, Yu et al. proposed two sphere-based PSNR computations, named
S-PSNR and L-PSNR respectively. 
The S-PSNR weights all position on the sphere equally while the 
L-PSNR weights positions according to their latitudes and access frequency. 
In this paper, we use these two metrics to measure the quality of encoded video 
in Section \ref{sec:exp} and used L-PSNR to replace regular PSNR in the BD-Rate calculations.

\section{Tile segmentation scheme}
\label{sec:tile}

We start by analyzing the equirectangular projection and striving to minimize the pixel area after the projection.
We divide a sphere into several pieces along latitude lines into two camber surfaces and rings,
flatten the camber surfaces into circles and cut open the rings, and then unfold and reshape them into rectangular tiles.
Finally, we expand the two circles corresponding to the north and south poles of the original sphere 
to their bounding square with a black filling, as shown in Fig.\ref{fig:seg}(a).
The resulted tiles are still approximately equal in size. 

Following these steps, with the number of titles increasing, the afore-described mapping 
will create a set of approximately equal-area titles with the total area approaching the that of the 
original sphere.
\subsection{Area comparison}
\label{sec:tile_area}
The total area of each hemisphere's tiles can be describe as:
\begin{equation}
\label{eq:area}
{S}_{hemisphere} = {S}_{pole} + \sum_{i=1}^{n} 2\pi {r}^{2}\cos{{\theta}_{i-1}}({\theta}_{i}-{\theta}_{i-1}) ~.
\end{equation}
\begin{equation}
\label{eq:pole}
{S}_{pole}=
\left\{\begin{matrix}
\pi{(\frac{\pi}{2} - {\theta}_{p})}^{2}{r}^{2}&, \text{Circle Pole} \\ 
4{(\frac{\pi}{2} - {\theta}_{p})}^{2}{r}^{2} &, \text{Square Pole}
\end{matrix}\right. ~.
\end{equation}
where ${\theta}_{1},{\theta}_{2}...{\theta}_{n} $ are the latitude (in rad) of the borders, 
e.g. for north hemisphere ${\theta}_{1},{\theta}_{2}...{\theta}_{n} \in (0,\frac{\pi}{2})$ and 
$ {\theta}_{1} < {\theta}_{2}<...<{\theta}_{n}={\theta}_{p} $.  
${\theta}_{0} = 0$ corresponds to the latitude of the equator, while ${\theta}_{p}$ is 
the latitude of the outer edge of the pole areas which is actually the same with ${\theta}_{n}$.
After $n$ cuts at each hemisphere, the sphere is projected and divided into $(2n+1)$ tiles,
with one big tile across the equator.

In the tile segmentation method proposed in Yu's work\cite{yu2015content}, 
the area of pole under same minimum sample density is 
\begin{equation}
\label{eq:poleyu}
{S}_{pole, \text{Yu}}=2\pi {r}^{2}(\frac{\pi}{2}-{\theta}_{p})\cos{{\theta}_{p}} ~.
\end{equation}
 which is always larger than Eq.\ref{eq:pole} .

\begin{figure}[htb]
\begin{minipage}[b]{1.0\linewidth}
  \centering
  \centerline{\includegraphics[width=8cm]{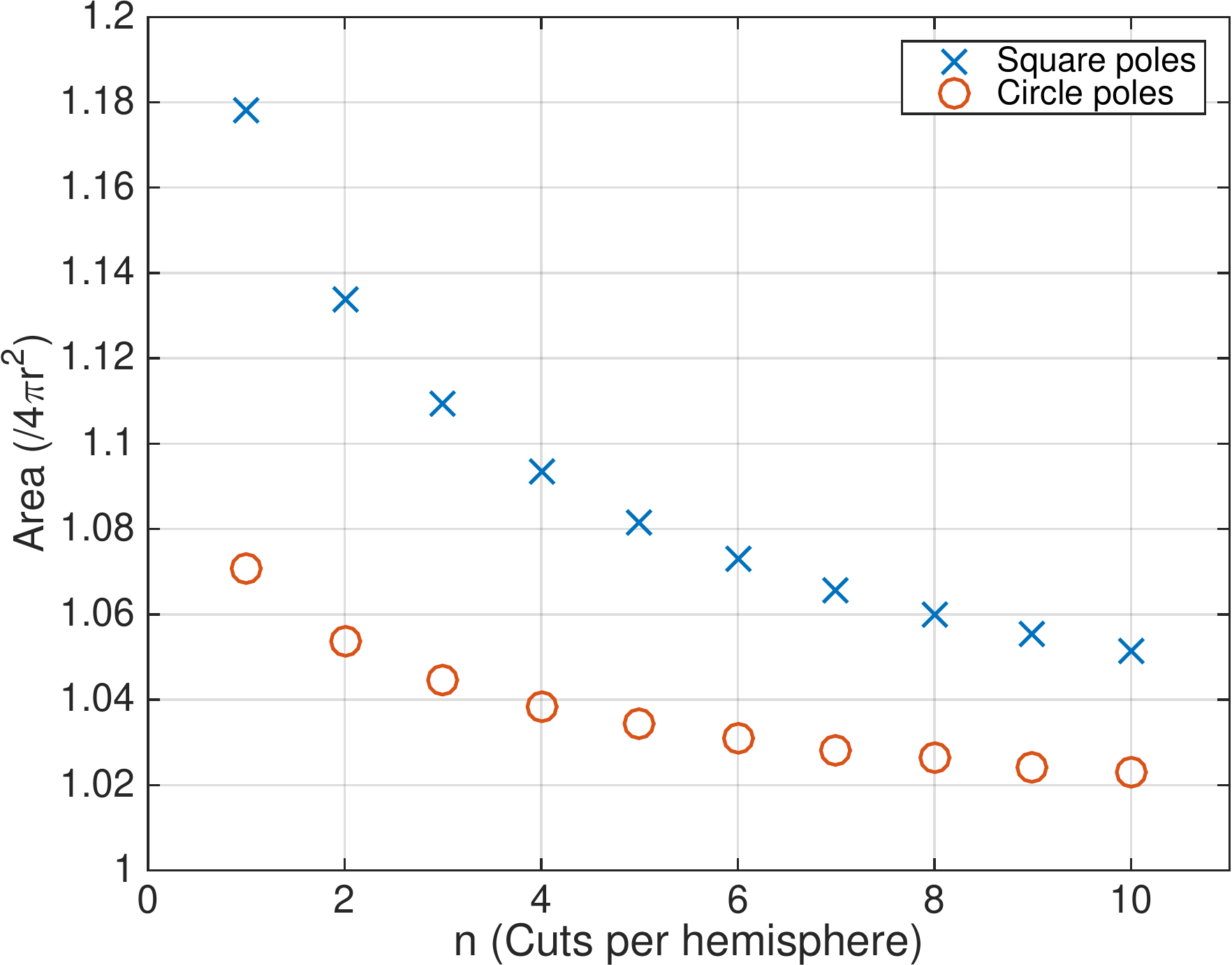}}
\end{minipage}
\caption{Relation between cut number and total area without overlap}
\label{fig:number}
\end{figure}

The best segmentation scheme can be found by minimizing the total area given a number of the tiles $n$:
\begin{equation}
\label{eq:optimize}
{\Theta}_{n} = \{{\theta}_{1},...,{\theta}_{n}\} = \argmin_{0<{\theta}_{1} < ...<{\theta}_{n}<\frac{\pi}{2}}{S}_{hemisphere}  ~.
\end{equation}
Some of the best schemes are shown in Fig.\ref{fig:number} and Table \ref{tab:opt}, where the results with the 
pole areas calculated as circles or black-filled to bounding boxes (\ref{eq:pole}) are marked
with circles and \textit{Circle Poles}, or crosses and
\textit{Square Poles}. The horizontal axis in the figure is the number of cuts
in each hemisphere, while the vertical axis shows the area as a ratio with regard to the area of the
original sphere.
\begin{figure}[htb]
\begin{minipage}[b]{1.0\linewidth}
  \centering
  \centerline{\includegraphics[width=8cm]{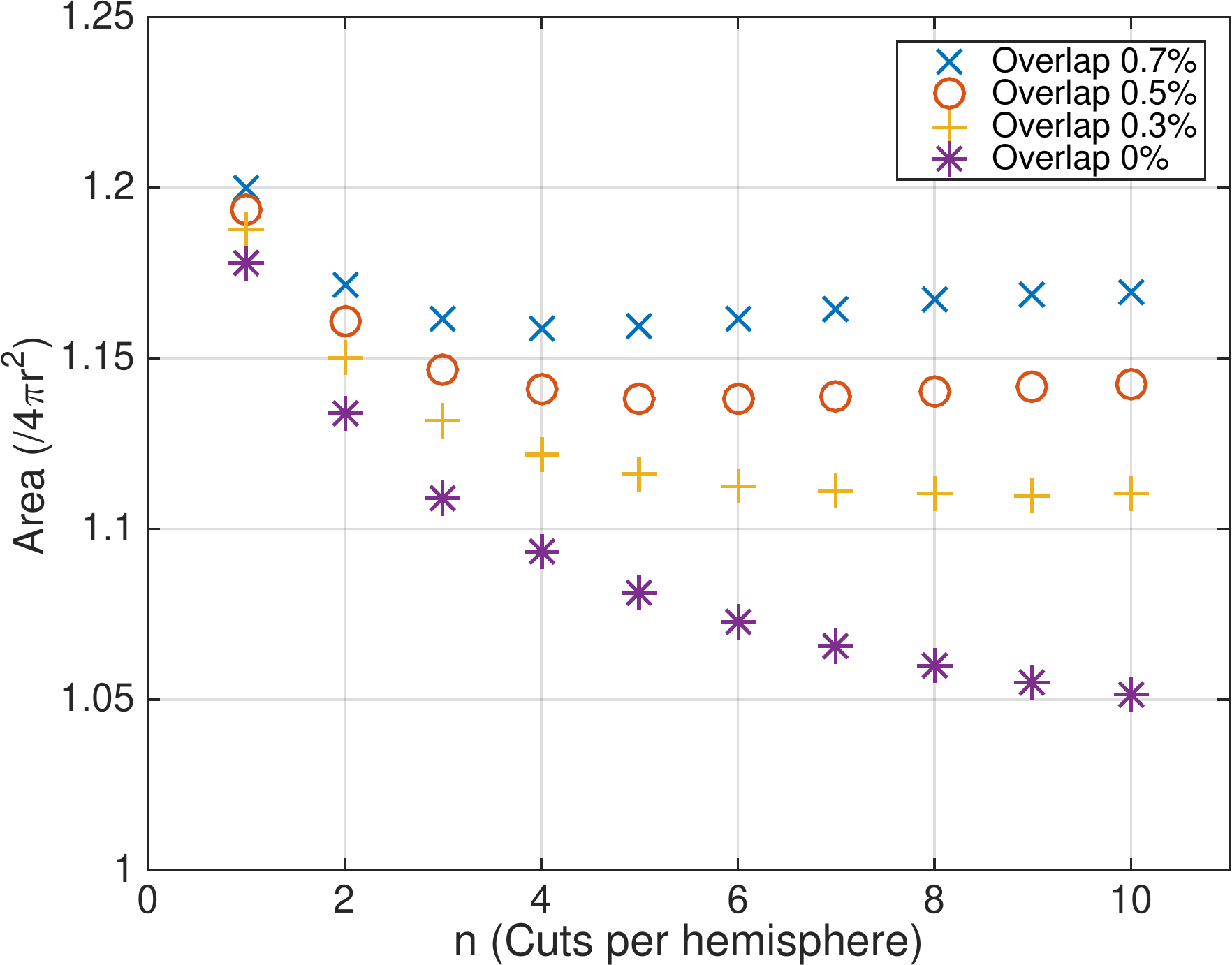}}
\end{minipage}
\caption{Total area comparison of different tile number and overlap (Square poles)}
\label{fig:overlap}
\end{figure}

Fig.\ref{fig:number} shows that increasing the number of tiles can effectively reduce the 
total area. It is of interest to note that when the number of tiles becomes very large, 
the corresponding tile based segmentation becomes the equal-area sinusoidal projection.

In practice however, the number of tiles is limited by the overhead introduced by using tile-based
video coding (when each tile is encoded independently) as well as the capability of the 
decoder to decode video encoded with multiple titles.

\subsection{Overlapping}
\label{sec:tile_overlapping}
When a tiled video is encoded at a low bitrate, border artifacts exist between
tile boundaries. A common approach to this issue is to set
overlapping areas between tiles. 

Introducing overlaps will increase the total area. Hence the 
optimization problem in (\ref{eq:optimize}) becomes:

\begin{equation}
\label{eq:overlap_area}
\begin{multlined}
{S}_{hemisphere}(\sigma) = {S}_{pole}(\sigma) + 2\pi {r}^{2}({\theta}_1 + \frac{\sigma\pi}{2}) \\
+ \sum_{i=2}^{n} 2\pi {r}^{2}\cos{{\theta}_{i-1}}({\theta}_{i}-{\theta}_{i-1} + \sigma\pi) ~.
\end{multlined}
\end{equation}

\begin{equation}
\begin{multlined}
\label{eq:overlap_pole}
{S}_{pole}(\sigma)=
\left\{\begin{matrix}
\pi{(\frac{\pi}{2} - {\theta}_{p} + \frac{\sigma\pi}{2})}^{2}{r}^{2}&, \text{Circle Pole} \\
4{(\frac{\pi}{2} - {\theta}_{p} + \frac{\sigma\pi}{2})}^{2}{r}^{2}&, \text{Square Pole}
\end{matrix}\right. ~.
\end{multlined}
\end{equation}

\begin{equation}
\label{eq:overlap_opt}
\begin{split}
{\Theta}_{n,\sigma} &= \{{\theta}_{1},...,{\theta}_{n}\} \\
&= \argmin_{\frac{\sigma\pi}{2}<{\theta}_{1} < ...<{\theta}_{n}<\frac{\pi-\sigma\pi}{2}}{S}_{hemisphere} (\sigma) ~.
\end{split}
\end{equation}

\begin{equation}
\label{eq:overlap_opt2}
{\Theta}_{\sigma} = \argmin_{{\Theta}_{n,\sigma} } {S}_{hemisphere} (\sigma) ~.
\end{equation}

where  ${\Theta}_{\sigma}$ is the optimization result, and $\sigma$ is the percentage of the overlap in height, e.g. 0.5\% overlap of a 1080p video is 5 pixel width.
Some of the solutions to the optimization problem are shown in Fig.\ref{fig:overlap} and Table \ref{tab:opt}.
For the same overall area, the increase of the overlapped areas will reduce the number of tiles,
from 41-tiles at 0.1\% overlap to 9-tiles at 0.7\% overlap.

\begin{table}[]
\centering
\caption{Optimization results on specified tiles number and overlap percentage (in degrees)}
\label{tab:opt}
\begin{tabular}{c|ccc}
\hline
Scheme                                                         & 0\%  overlap                                                            & 0.3\% overlap                                                             & 0.5\%  overlap                                                            \\ \hline
\begin{tabular}[c]{@{}c@{}}3-tiles\\ (Circle pole)\end{tabular} & ${\theta}_{1}$=32.70$^{\circ}$                                                        & ${\theta}_{1}$=32.97$^{\circ}$                                                      & ${\theta}_{1}$=33.15$^{\circ}$                                                         \\
\begin{tabular}[c]{@{}c@{}}5-tiles\\ (Circle pole)\end{tabular} & \begin{tabular}[c]{@{}c@{}}${\theta}_{1}$ =25.34$^{\circ}$\\  ${\theta}_{2}$ =38.22$^{\circ}$\end{tabular} & \begin{tabular}[c]{@{}c@{}}${\theta}_{1}$ =26.08$^{\circ}$\\ ${\theta}_{2}$ =38.81$^{\circ}$\end{tabular} & \begin{tabular}[c]{@{}c@{}}${\theta}_{1}$ =26.58$^{\circ}$ \\ ${\theta}_{2}$ =39.21$^{\circ}$\end{tabular} \\
\begin{tabular}[c]{@{}c@{}}3-tiles\\ (Square pole)\end{tabular} & ${\theta}_{1}$=45.00$^{\circ}$                                                            & ${\theta}_{1}$=45.27$^{\circ}$                                                        & ${\theta}_{1}$=45.45$^{\circ}$                                                         \\
\begin{tabular}[c]{@{}c@{}}5-tiles\\ (Square pole)\end{tabular} & \begin{tabular}[c]{@{}c@{}}${\theta}_{1}$ =35.07$^{\circ}$ \\ ${\theta}_{2}$ =53.17$^{\circ}$\end{tabular} & \begin{tabular}[c]{@{}c@{}}${\theta}_{1}$ =35.81$^{\circ}$\\ ${\theta}_{2}$ =53.77$^{\circ}$\end{tabular} & \begin{tabular}[c]{@{}c@{}}${\theta}_{1}$ =36.30$^{\circ}$ \\ ${\theta}_{2}$ =54.18$^{\circ}$\end{tabular} \\ \hline
\end{tabular}
\end{table}

\subsection{Rearrangement and layout}
When some decoders can not decode multiple streams at the same time, compacting all 
the tiles into one rectangular video stream is needed for compatibility. 

Fig. \ref{fig:seg} (b) and (c) are examples of such layout transforming
 multiple tiles into a single rectangular tile.
As the number of tiles increases, more pixels will be wasted due to layout overhead.
It's worth noting that
the layout in Fig.\ref{fig:seg} (b) with segmentation at $\theta_{1}=45^{\circ}$ 
is the optimum for 3-tiles, from both layout and sample density 
(shown in Table \ref{tab:opt}) aspect.

\section{Experimental results}
\label{sec:exp}

\subsection{Non-overlapping RD performance}
The RD performance experiments were conducted with 
the widely used open source HEVC encoder x265\cite{x265}
on its default settings. The test sequences are from two datasets. 
The first 10 sequences in Table \ref{tab:s_bd} and Table \ref{tab:l_bd} are 9K images from SUN360 database\cite{xiao2012recognizing},
and the last two sequences are 4K video clips generously provided by LETINVR.

The QPs used for computing the BD-rate are 22, 27, 32, 37.
For the quality metric, we use the S-PSNR and L-PSNR as in \cite{yu2015framework} 
and introduced in Section \ref{sec:r_work}.
The S-PSNR weights all position on the sphere equally while the L-PSNR weights positions 
according to its latitude access frequency. In our experiments, we measured both metrics
and we used both S-PSNR and L-PSNR in the BD-Rate calculations. 

As the test sequences are all in equirectangular projection format as ground truth, 
we down-sampled the sequences in the experiments so that the ground truth has always higher 
sample density. For example, the SUN360 experiments were conducted at 4552x2276
while the inputs are at 9104x4552 equirectangular.
When calculating S-PSNR and L-PSNR, decoded sequences were 
mapped back to equirectangular with 
the original input resolution.
Bilinear interpolation was used in above steps.

\begin{figure*}[htb]
\begin{minipage}[b]{0.475\linewidth}
  \centering
  \centerline{\includegraphics[width=6.0cm]{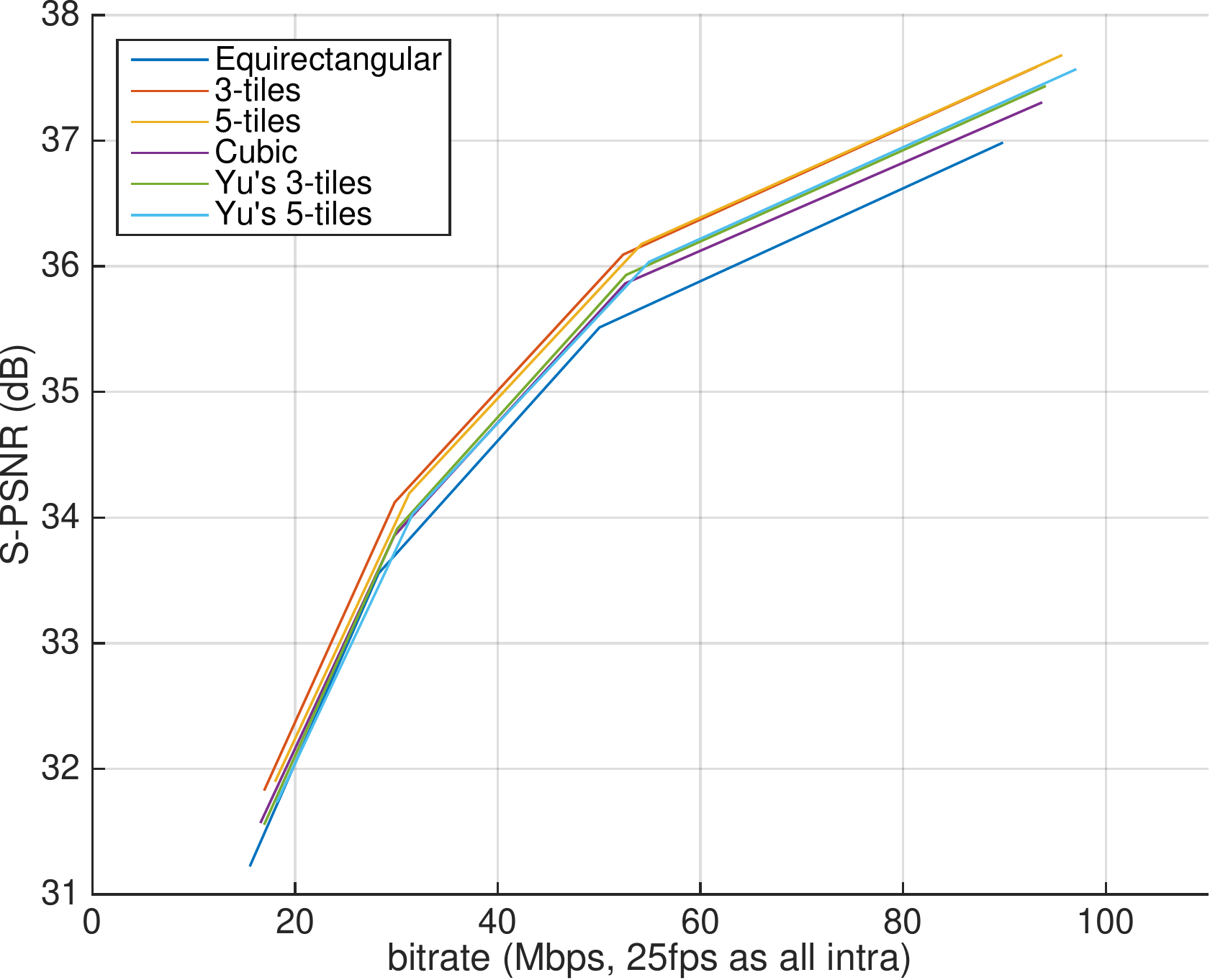}}
  \centerline{(a) RD curve of Seq. 1 using S-PSNR}\medskip
\end{minipage}
\hfill
\begin{minipage}[b]{0.475\linewidth}
  \centering
  \centerline{\includegraphics[width=6.0cm]{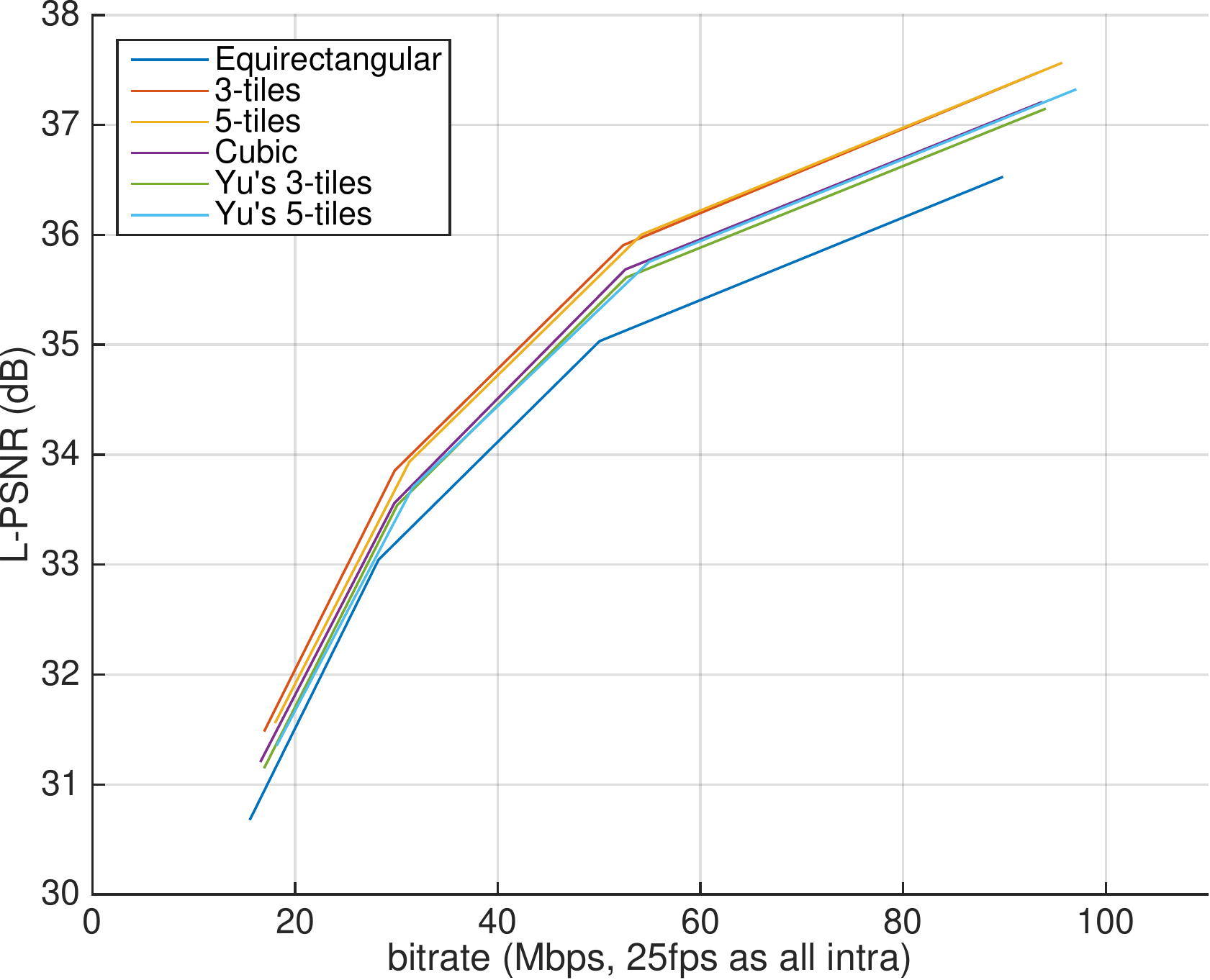}}
  \centerline{(b) RD curve of Seq. 1 using L-PSNR}\medskip
\end{minipage}
%
\caption{RD curves of two sequences with S-PSNR and L-PSNR metric. }
\label{fig:rd}
\end{figure*}

The experiment results
are shown in Fig.\ref{fig:rd}, Table \ref{tab:s_bd} and Table \ref{tab:l_bd},
including proposed 3-tiles scheme (P3), proposed 
5-tiles scheme (P5), cubic projection (CP). 
In \cite{yu2015content}, Yu et al. proposed a tile 
segmentation method and here we implement it as 3-tiles (Y3) and 5-tiles (Y5).

Table \ref{tab:s_bd} and \ref{tab:l_bd} show about 12\% and 20\%
 bitrate savings to equirectangular projection
 on S-PSNR and L-PSNR respectively, 
 which are 4\% and 6\% more compared to the previous method.

\begin{table}[]
\centering
\caption{BD-Rate of proposed method using S-PSNR}
\label{tab:s_bd}
\begin{tabular}{r|rrrrr}
\hline
\multicolumn{1}{c|}{Seq.} & \multicolumn{1}{c}{P3} & \multicolumn{1}{c}{P5} & \multicolumn{1}{c}{CP} & \multicolumn{1}{c}{Y3} & \multicolumn{1}{c}{Y5} \\ \hline
1 & -11.6\% &   -9.8\% &  -4.6\% &  -5.5\% &  -4.5\% \\
2 & -13.1\% &   -13.4\% &   -9.2\% &  -10.2\% &   -10.4\% \\
3 & -12.5\% &   -11.8\% &   -12.2\% &   -7.4\% &  -7.5\% \\
4 & -10.6\% &   -10.0\% &   -7.3\% &  -6.8\% &  -7.2\% \\
5 & -15.2\% &   -16.4\% &   -10.3\% &   -12.3\% &   -12.9\% \\
6 & -8.0\% &  -14.7\% &   -3.1\% &  -7.5\% &  -7.7\% \\
7 & -12.9\% &   -13.1\% &   -8.2\% &  -10.3\% &   -11.0\% \\
8 & -8.4\% &  -7.0\% &  -1.4\% &  -4.5\% &  -3.9\% \\
9 & -1.4\% &  1.0\% &   1.2\% &   0.9\% &   2.8\% \\
10 & -13.7\% &   -13.4\% &   -9.6\% &    -9.7\% &  -10.9\% \\
11 & -16.8\% &   -18.6\% &   -0.4\% &    -11.3\% &   -14.6\% \\
12 & -22.5\% &   -23.9\% &   -7.8\% &    -13.6\% &   -17.5\%  \\ \hline
\multicolumn{1}{l|}{Avg.} & -12.2\% &    -12.6\% &   -6.1\% &  -8.2\% &  -8.8\% \\ \hline
\end{tabular}
\centering
\caption{BD-Rate of proposed method using L-PSNR}
\label{tab:l_bd}
\begin{tabular}{r|rrrrr}
\hline
\multicolumn{1}{c|}{Seq.} & \multicolumn{1}{c}{P3} & \multicolumn{1}{c}{P5} & \multicolumn{1}{c}{CP} & \multicolumn{1}{c}{Y3} & \multicolumn{1}{c}{Y5} \\ \hline
1                         & -18.5\%                & -17.5\%                & -11.8\%                & -10.0\%                & -10.3\%                \\
2                         & -17.9\%                & -18.3\%                & -12.3\%                & -13.6\%                & -14.7\%                \\
3                         & -19.7\%                & -19.2\%                & -18.8\%                & -12.2\%                & -13.2\%                \\
4                         & -19.8\%                & -19.3\%                & -17.3\%                & -11.3\%                & -12.9\%                \\
5                         & -24.6\%                & -25.0\%                & -22.1\%                & -16.2\%                & -17.8\%                \\
6                         & -20.4\%                & -19.9\%                & -18.7\%                & -12.7\%                & -14.4\%                \\
7                         & -27.6\%                & -27.0\%                & -24.7\%                & -16.9\%                & -19.5\%                \\
8                         & -16.4\%                & -15.5\%                & -9.7\%                 & -8.5\%                 & -9.0\%                 \\
9                         & -10.6\%                & -8.3\%                 & -7.4\%                 & -4.2\%                 & -3.6\%                 \\
10                        & -21.7\%                & -21.3\%                & -10.0\%                & -13.8\%                & -16.2\%                \\
11                        & -22.5\%                & -23.9\%                & -7.8\%                 & -24.1\%                & -23.9\%                \\
12                        & -21.9\%                & -21.0\%                & -5.0\%                 & -11.9\%                & -13.1\%                \\ \hline
\multicolumn{1}{l|}{Avg.} & -20.1\%                & -19.7\%                & -14.6\%                & -13.0\%                & -14.1\%                \\ \hline
\end{tabular}
\centering
\caption{BD-rate result of 0.5\% overlap compare to 0\% overlap}
\label{tab:overlap}
\begin{tabular}{r|rr|rr}
\hline
\multicolumn{1}{c|}{\multirow{2}{*}{Seq.}} & \multicolumn{2}{c|}{S-PSNR}                      & \multicolumn{2}{c}{L-PSNR}                      \\ \cline{2-5} 
\multicolumn{1}{c|}{}                      & \multicolumn{1}{c}{P3} & \multicolumn{1}{c|}{P5} & \multicolumn{1}{c}{P3} & \multicolumn{1}{c}{P5} \\ \hline
1                                          & 0.8\%                      & 1.2\%                       & 0.9\%                      & 1.4\%                      \\
2                                          & 1.4\%                      & 0.4\%                       & 1.6\%                      & 0.4\%                      \\
3                                          & 0.9\%                      & 2.3\%                       & 0.8\%                      & 2.2\%                      \\
4                                          & 1.0\%                      & 1.4\%                       & 1.1\%                      & 1.6\%                      \\
5                                          & 1.0\%                      & 1.9\%                       & 1.1\%                      & 1.9\%                      \\
6                                          & 1.1\%                      & 2.3\%                       & 1.3\%                      & 2.6\%                      \\ \hline
Avg.                                       & 1.0\%                      & 1.6\%                       & 1.1\%                      & 1.7\%                      \\ \hline
\end{tabular}
\end{table}
\subsection{Overlapping de-blocking performance}
\begin{figure}[htb]
\begin{minipage}[b]{1.0\linewidth}
  \centering
  \centerline{\includegraphics[width=8cm]{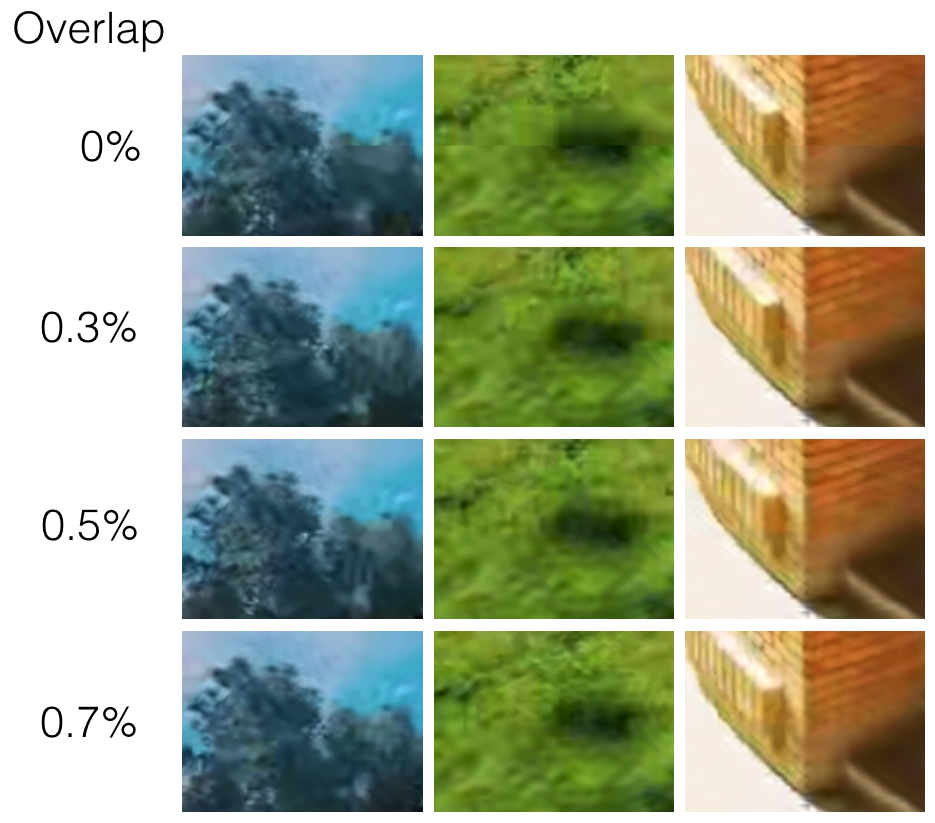}}
\end{minipage}
\caption{Comparison on the border of tile with 0\%, 0.3\%, 0.5\% and 0.7\% overlap from top to bottom, QP = 37.}
\label{fig:overlap_effect}
\end{figure}
In this section we compare the non-overlapped and different 
percentage overlapped result with 3-tiles and 5-tiles configuration.
The output sequence was encoded with QP 22, 27, 32, 37 
and different overlap configuration, then decoded and blended the overlapping area.

The third row in Fig.\ref{fig:overlap_effect}, shows 
the de-blocking result of 0.5\% overlap, which gives a well enough transition.
For a 4k video, 0.5\% overlap costs 10 pixel width.

The BD-rate results of proposed 3-tiles and 5-tiles scheme 
with 0.5\% overlap compared to non-overlap is shown in Table \ref{tab:overlap}, 
where such overlap scheme introduces no more than 1.7\% bitrate overhead averagely.

\section{conclusion}
\label{sec:conclusion}
In this paper, we proposed a tile segmentation scheme for omnidirectional video encoding,
which has less pixel wastage and more legitimate mapping method.
Experimental results show the proposed scheme can save up to 28\% of pixel area and 20\% BD-rate averagely compare to traditional equirectangular projection.

Future work includes optimization on segmentation scheme targeting on region of interest (ROI) application,
as well as optimizations on temporal and spacial random access.

\label{sec:ref}

\bibliographystyle{IEEEbib}
\bibliography{main}

\end{document}